\documentclass[runningheads]{llncs}
\usepackage[T1]{fontenc}
\usepackage{amssymb}
\usepackage{xcolor}
\usepackage{booktabs}
\usepackage{multirow}
\usepackage{caption}
\usepackage{subcaption}
\usepackage[dvipsnames,table]{xcolor}
\usepackage{array}
\usepackage{graphicx,verbatim}
\usepackage{amsmath}
\usepackage{amsfonts}
\usepackage{url}
\usepackage{marvosym}
\usepackage{xurl}
\usepackage[colorlinks=true, urlcolor=blue, breaklinks=true]{hyperref}

\begin{document}
\title{C\textsuperscript{2}RM-Seg: Causal Counterfactual Reasoning with Structural-Semantic Priors for Weakly Supervised Histopathological Tissue Segmentation}

\author{Hualong Zhang\inst{1} \and
Siyang Feng\inst{1} \and
Zihan Huan\inst{1} \and
Yi Qian \inst{1} \and
Zhenbing Liu\inst{1} \and
Rushi Lan\inst{1,2(\textrm{\Letter})} \and
Xipeng Pan\inst{1,3}
}
\authorrunning{H. Zhang et al.}
%
\institute{
Guangxi Key Laboratory of Image and Graphic Intelligent Processing, Guilin University of Electronic Technology, Guilin 541004, China
\and
International Joint Research Laboratory of Spatio-temporal Information and Intelligent Location Services, Guilin University of Electronic Technology, Guilin 541004, China
\and
School of Computer Science and Information Security, Guilin University of Electronic Technology, Guilin 541004, China
\\
\email{rslan@guet.edu.cn}
}
\titlerunning{C\textsuperscript{2}RM-Seg for Histopathological Tissue Segmentation}
\maketitle  
\begin{abstract}
Histopathological tissue segmentation is essential for computer-aided diagnosis, yet weakly supervised methods often suffer from noisy pseudo-labels generated by Class Activation Mapping (CAM). Existing CAM approaches tend to focus on staining-driven appearance cues rather than true causal tissue morphology, resulting in spurious localization and poor structural consistency. To address this issue, we propose C\textsuperscript{2}RM-Seg, a two-stage framework that integrates causal pseudo-label refinement with structure-aware semantic enhancement. For classification, we introduce a Causal Counterfactual Reasoning Module (C\textsuperscript{2}RM) that decomposes features into latent factors and performs counterfactual intervention via a learned causal structure matrix, suppressing confounding context and producing morphology-aligned CAMs. For segmentation, we design a Dual-Path Structural–Semantic Architecture that combines fine-grained structural features from ResNeSt with global semantic priors from frozen DINOV3 foundation model. A cross-path gating mechanism adaptively regulates semantic injection using local structural cues to preserve boundary fidelity. To further mitigate residual pseudo-label noise, we propose an Uncertainty-Gated Margin (UGM) loss, which dynamically balances margin enforcement and confidence learning based on prediction uncertainty. Extensive experiments on two public histopathological tissue datasets show that C\textsuperscript{2}RM-Seg achieves state-of-the-art performance. Code is available at \href{https://github.com/OceanPetal/C2AM-Seg}{\nolinkurl{https://github.com/OceanPetal/C2AM-Seg}}.


\end{abstract}
\section{Introduction}
\label{sec:intro}
Histopathological tissue segmentation plays an important role in computer-aided diagnosis, enabling quantitative analysis for cancer staging and prognosis assessment \cite{andani2025histopathology,wang2024pathology,feng2025weakly,an2025mlif}. Accurate delineation of pathological regions provides structured representations that facilitate downstream clinical interpretation \cite{tang2025prototype,cheng2025fcaformer,pan2025weakly}. However, fully supervised segmentation requires dense pixel-level annotations, which are costly and highly dependent on expert knowledge, especially for large and heterogeneous whole-slide images \cite{chikontwe2022weakly,li2023weakly}. Weakly Supervised Semantic Segmentation (WSSS) alleviates this burden by learning dense predictions from only image-level labels, offering a scalable alternative for histopathology analysis \cite{fan2025dipathmamba,zhang2025edge,huan2026bridging}.

Most WSSS methods adopt Class Activation Mapping (CAM) to generate coarse seeds and then train a segmentation model with the resulting pseudo-masks \cite{kang2025exploring,fu2025cam,feng2026long,feng2026qupas}. Recent efforts refine CAMs through activation enhancement and uncertainty modeling. Representative approaches include Wave-aware CAM \cite{feng2025wave}, UAM \cite{kang2025exploring}, and UM-CAM \cite{fu2025cam}, as well as histopathology-oriented variants such as ARML \cite{feng2024mining} and HAMIL \cite{zhong2023hamil}. Despite improved localization, existing CAM-based refinements often conflate causal morphology with confounded cues (e.g., stain variations and background tissue patterns). Without explicitly modeling latent confounders or performing causal intervention, CAM responses can be systematically skewed toward context, leading to biased pseudo-masks. Beyond CAM refinement, structural modeling strategies such as MLPS \cite{han2022multi}, ESFAN \cite{zhang2025edge}, and expert-guided multi-granularity fusion \cite{hu2026expert} aim to better preserve fine-grained morphology and boundary consistency. Noise-aware optimization techniques like OEEM \cite{li2022online} further attempt to mitigate pseudo-label noise. However, when pseudo-labels are systematically biased by uncontrolled confounding factors, architectural refinement or loss reweighting alone may be insufficient to resolve the underlying causal misalignment.

To address these challenges, we propose \textbf{C\textsuperscript{2}RM-Seg}, a framework that combines causal deconfounding with foundation model priors for robust weakly supervised histopathology segmentation. We introduce a Causal Counterfactual Reasoning Module (C\textsuperscript{2}RM) that formulates CAM generation under a Structural Causal Model (SCM) and performs counterfactual intervention to subtract confounding effects, producing cleaner seeds. We further design a Dual-Path Structural--Semantic Architecture that integrates a ResNeSt \cite{zhang2022resnest} branch for local morphology and a frozen DINOV3 \cite{simeoni2025dinov3} branch for global semantic priors, coupled with a cross-path gating mechanism for adaptive fusion. Finally, an Uncertainty-Gated Margin (UGM) loss enables noise-tolerant optimization from imperfect pseudo-labels. Our main contributions are summarized as follows:
\begin{itemize}
\item We identify that CAM-based methods are biased toward staining-driven appearance rather than causal tissue morphology, and propose C\textsuperscript{2}RM, a counterfactual causal module that removes confounding context to produce high-fidelity pseudo-labels.
\item We propose a Dual-Path Structural--Semantic segmentation framework where morphological structure guides foundation semantic injection via cross-path gating, integrating ResNeSt with a frozen DINOV3 for boundary-preserving global semantics, together with an Uncertainty-Gated Margin (UGM) loss for noise-tolerant optimization.
\item Extensive experiments on public histopathology benchmarks demonstrate that C\textsuperscript{2}RM-Seg outperforms state-of-the-art WSSS methods in both segmentation accuracy and boundary quality.
\end{itemize}
\section{Method}
\subsection{Overview}
\begin{figure}[t]
    \centering
    \includegraphics[width=\textwidth]{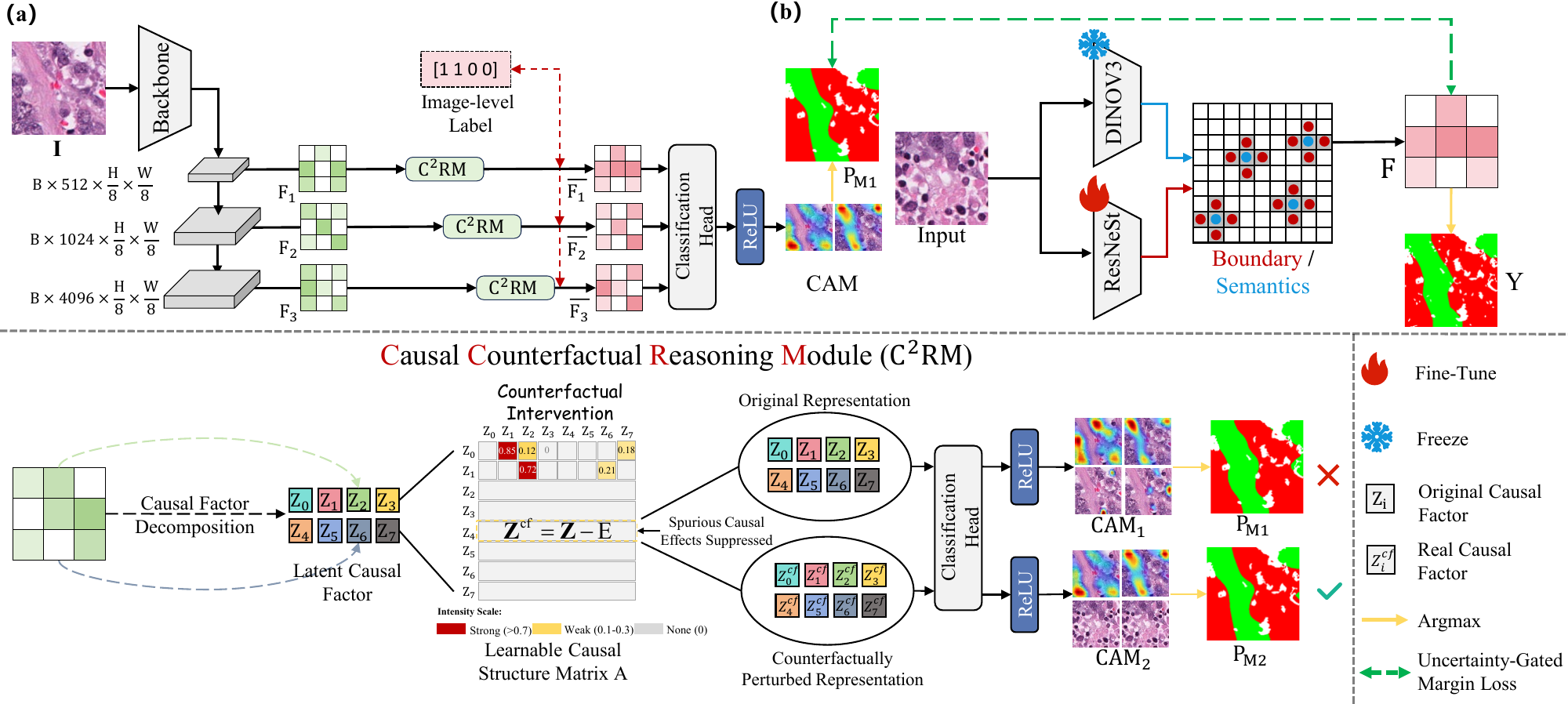}
    \caption{Overview of C\textsuperscript{2}RM-Seg. (a) Pseudo-Mask Generation: The C\textsuperscript{2}RM rectifies CAMs by performing counterfactual intervention $Z^{cf} = Z - E$ on latent factors $Z$ using a learnable causal matrix $A$. (b) Segmentation Training: A Dual-Path Structural-Semantic Architecture synergizes ResNeSt \cite{zhang2022resnest} features with frozen DINOV3 \cite{simeoni2025dinov3} priors via cross-path gating. The model is optimized using a UGM loss that dynamically re-weights supervision based on prediction uncertainty to mitigate label noise.}
\label{fig:framework}
\end{figure}

As shown in Fig.~\ref{fig:framework}, given a histopathology image $I$ with an image-level label $y$, C\textsuperscript{2}RM-Seg adopts a two-stage weakly supervised pipeline. First, we generate pseudo-masks by producing CAMs with the proposed C\textsuperscript{2}RM, which reduces spurious contextual bias via counterfactual de-confounding. Second, we train a segmentation network with a Dual-Path Structural--Semantic Architecture that fuses local morphological cues and global foundation priors through cross-path gating. Finally, a UGM loss is used to alleviate residual noise in pseudo-masks during optimization.

\subsection{Causal Counterfactual Reasoning Module}
Spurious correlations are prevalent in histopathological images, where label-relevant tissue morphology, such as tumor, necrotic, lymphocyte-rich, and stromal regions, is often entangled with label-irrelevant acquisition and style factors, including staining variability, scanner-specific responses, tissue-processing artifacts, and co-occurring background textures. We regard these style-related variations as latent confounders and introduce C$^2$RM, an SCM-inspired counterfactual reasoning module that mitigates their influence by performing de-confounding through counterfactual feature correction.

Let $X\in\mathbb{R}^{C\times H\times W}$ be encoder features. We assume latent morphology causes $M$ and latent style/acquisition variables $S$ jointly affect the learned factors $\mathbf{Z}$, and $\mathbf{Z}$ determines the prediction $\hat{Y}$ (i.e., $S\rightarrow \mathbf{Z}\rightarrow \hat{Y}$ and $M\rightarrow \mathbf{Z}\rightarrow \hat{Y}$). Our goal is to estimate counterfactual factors where incoming confounding influences are removed.

We project channels into $K$ factor subspaces ($D=C/K$) with a learnable $1\times1$ projection $\Phi(\cdot)$ and compute global factor statistics by average pooling:
\begin{equation}
\mathbf{z}_k=\frac{1}{HW}\sum_{h=1}^{H}\sum_{w=1}^{W}\Phi_k(X_{h,w}),\quad
k\in\{1,\dots,K\},\ \mathbf{z}_k\in\mathbb{R}^{D}.
\label{eq:z_pool_short}
\end{equation}

We learn a directed factor graph with adjacency $A_c\in\mathbb{R}^{K\times K}$, where $A_{c,i,j}$ represents the influence of factor $j$ on factor $i$. To avoid self-loops, we mask the diagonal and apply row-wise softmax:
\begin{equation}
\mathbf{A}=\text{softmax}\!\big(A_c\odot(\mathbf{1}-I)\big),\quad A_{i,i}=0.
\label{eq:A_short}
\end{equation}
The confounding contribution to factor $i$ is estimated by aggregating its parents:
\begin{equation}
\mathbf{E}_i=\sum_{j\neq i}A_{i,j}\mathbf{z}_j.
\label{eq:E_short}
\end{equation}

Under a linear additive SCM approximation for factor interactions, removing incoming confounding edges yields a counterfactual factor that subtracts the estimated parental contribution:
\begin{equation}
\mathbf{z}_i^{cf}\approx \mathbf{z}_i-\mathbf{E}_i,\qquad
\mathbf{Z}^{cf}=\mathbf{Z}-\mathbf{A}\mathbf{Z},
\label{eq:zcf_short}
\end{equation}
where $\mathbf{Z}\in\mathbb{R}^{K\times D}$ stacks $\{\mathbf{z}_k\}_{k=1}^{K}$. This operation suppresses components explainable by co-varying style/background factors (e.g., stain-driven textures) and preserves morphology-dominant signals.

Finally, we inject $\mathbf{Z}^{cf}$ back to recalibrate spatial features via a gated residual correction:
\begin{equation}
\mathbf{G}=\sigma\!\left(\mathcal{R}\big(\text{Broadcast}(\mathbf{Z}^{cf})\big)\right),\qquad
\hat{X}=X+\gamma\cdot(\mathbf{G}\odot X),
\label{eq:refine_short}
\end{equation}
where $\mathcal{R}(\cdot)$ is a $1\times1$ bottleneck block, $\sigma$ is Sigmoid, $\text{Broadcast}(\mathbf{Z}^{cf}) \in \mathbb{R}^{C \times H \times W}$ 
replicates the channel-wise counterfactual factors along spatial dimensions, and $\gamma$ is a learnable scalar.

\subsection{Dual-Path Structural-Semantic Architecture}
To bridge the gap between local morphological fidelity and global pathological semantics, we propose a Dual-Path Structural-Semantic Architecture. This framework is designed to synergize high-resolution structural textures with robust features from large-scale foundation models, while simultaneously addressing the label noise inherent in pseudo-mask supervision.

The architecture comprises two specialized branches. The Structural Branch utilizes a ResNeSt-200e backbone followed by a pyramid pooling module to extract multi-scale spatial features, denoted as $F_{S} \in \mathbb{R}^{C \times H \times W}$. In parallel, the semantic branch leverages a vision transformer (ViT-B/16) pre-trained via the DINOV3 objective. To exploit the generalized representations of the foundation model while preventing overfitting to noisy pseudo-labels, the ViT backbone parameters $\Theta_{ViT}$ remain frozen. A learnable semantic adapter $\mathcal{A}_c$ projects the ViT embeddings into the structural manifold:
\begin{equation}
    F_{V} = \text{Up} \left( \sigma ( \mathcal{B} ( \mathbf{W}_{ad} \cdot \mathcal{E}_{ViT}(X; \Theta_{ViT}) ) ) \right),
\end{equation}
where $\mathcal{E}_{ViT}$ represents the frozen encoder, $\mathbf{W}_{ad}$ denotes the adapter weights, and $\text{Up}(\cdot)$ denotes bilinear upsampling for spatial alignment. 

To integrate these heterogeneous representations, we employ a cross-path gating mechanism. We posit that global semantic priors should be modulated by local structural evidence to maintain boundary integrity. A spatial gating map $G = \sigma(\Psi(F_S))$ is derived from $F_S$ to selectively weight the semantic contribution:
\begin{equation}
    \hat{F} = F_S + \sigma(\Psi(F_S)) \odot F_V,
\end{equation}
where $\Psi$ denotes a $1 \times 1$ convolutional bottleneck and $\odot$ represents the element-wise Hadamard product. This gating ensures that semantic information is concentrated on regions with high structural saliency.

Supervision via pseudo-masks introduces aleatoric uncertainty, particularly at tissue interfaces. To robustify the learning process, we formulate the Uncertainty-Gated Margin (UGM) loss. For each pixel $i$, let $p_{i,y}$ be the predicted probability for the target class $y$ and $p_{i,s} = \max_{j \neq y} p_{i,j}$ be the probability of the most competitive alternative class. We define the pixel-wise margin violation $\mathcal{M}_i$ and the confidence penalty $\mathcal{C}_i$ as:
\begin{equation}
    \mathcal{M}_i = [m - (p_{i,y} - p_{i,s})]_+, \quad \mathcal{C}_i = -\log(p_{i,y} + \epsilon),
\end{equation}
where $m$ is a predefined safety margin, $[\cdot]_+$ denotes the ReLU operation, and $\epsilon$ is a small constant for numerical stability. 

The total optimization objective is governed by a dynamic uncertainty weight $\alpha_i$, which adaptively shifts the focus between margin enforcement and confidence maximization. Given the local uncertainty $u_i = 1 - p_{i,y}$, the integrated loss function $\mathcal{L}$ for a set of valid pixels $\Omega$ is defined as:
\begin{equation}
    \mathcal{L} = \frac{1}{|\Omega|} \sum_{i \in \Omega} \left[ \frac{1}{1+e^{-\kappa(u_i - 0.5)}} \mathcal{C}_i + \left( 1 - \frac{1}{1+e^{-\kappa(u_i - 0.5)}} \right) \mathcal{M}_i \right],
\end{equation}
where $\kappa$ is a scaling factor controlling the transition steepness. This formulation ensures that for certain predictions (low $u_i$), the model prioritizes expanding the decision boundary margin, whereas for uncertain or potentially mislabeled regions (high $u_i$), it reverts to a conservative confidence-based penalty to avoid over-penalizing ambiguous samples.
\vspace{-3mm}
\section{Experiments}
\subsection{Experimental Setup} \textbf{Datasets.} We evaluate C\textsuperscript{2}RM-Seg on two public datasets. \textbf{LUAD-HistoSeg}~\cite{han2022multi} (lung adenocarcinoma) comprises 54 WSIs with four phenotypes (TE, TAS, NEC, LYM). It contains 16,678 training patches (image-level labels) and 607 validation/testing patches with dense pixel-level annotations. \textbf{BCSS}~\cite{amgad2019structured} (breast cancer) includes 151 WSIs targeting TUM, STR, LYM, and NEC, split into 23,422 training patches and 8,404 validation/testing patches with ground-truth masks. 

\noindent\textbf{Implementation Details.} Experiments are conducted in PyTorch on a NVIDIA RTX 3090. The classification stage uses ResNet38d with $K{=}8$ latent factors, trained for 20 epochs (batch size 20) via PolyOptimizer (initial LR $1{\times}10^{-2}$). For segmentation, the structural branch (ResNeSt-200e) is optimized by SGD (LR $1{\times}10^{-2}$), while the semantic branch (frozen DINOV3-Base + adapter) is trained via AdamW (LR $5{\times}10^{-3}$). We set the loss hyperparameters to $m{=}0.3$ and $\kappa{=}5$. Performance is reported using mIoU, bIoU, and HD95.
\begin{table*}[!t]
\centering
\caption{Quantitative comparison on the BCSS-WSSS and LUAD-HistoSeg datasets. The best and second-best results are highlighted in \textbf{bold} and underlined, respectively. The reported standard deviations are computed over five independent runs.}
\label{tab:comparison_subset}
\small
\resizebox{\textwidth}{!}{
\begin{tabular}{clc cc ccc ccc}
\toprule
\multirow{2}{*}{Sup.} & \multirow{2}{*}{Method} & \multirow{2}{*}{Venue} &
\multicolumn{2}{c}{Efficiency} &
\multicolumn{3}{c}{BCSS-WSSS} &
\multicolumn{3}{c}{LUAD-HistoSeg} \\
\cmidrule(lr){4-5}
\cmidrule(lr){6-8}
\cmidrule(lr){9-11}
& & &
GFLOPs$\downarrow$ & FPS$\uparrow$ &
mIoU$\uparrow$ & bIoU$\uparrow$ & HD95$\downarrow$ &
mIoU$\uparrow$ & bIoU$\uparrow$ & HD95$\downarrow$ \\
\midrule
\multirow{3}{*}{$\mathcal{F}$}
& SAM \cite{kirillov2023segment} & ICCV'23 & 5471.62 & 2.11 &
66.35$_{0.78}$ & 44.26$_{0.46}$ & 33.45$_{0.82}$ &
69.37$_{0.37}$ & 48.38$_{0.26}$ & 24.81$_{0.63}$ \\
& SAM2 \cite{ravi2024sam} & arXiv'24 & 121.68 & 135.45 &
69.44$_{0.93}$ & \underline{48.91$_{0.78}$} & 29.87$_{0.75}$ &
77.02$_{0.29}$ & 45.77$_{0.13}$ & 20.35$_{0.48}$ \\
& MedSAM \cite{ma2024segment} & NC'24 & 738.01 & 9.59 &
65.23$_{0.61}$ & 48.07$_{0.52}$ & 34.12$_{1.04}$ &
75.23$_{0.51}$ & 43.22$_{0.38}$ & 21.64$_{0.72}$ \\
\midrule
\multirow{8}{*}{$\mathcal{W}$}
& TPRO \cite{zhang2023tpro} & MICCAI'23 & \textbf{67.79} & \textbf{203.34} &
68.17$_{0.83}$ & 45.84$_{0.47}$ & 31.56$_{0.94}$ &
76.53$_{0.92}$ & 46.67$_{0.41}$ & 20.92$_{0.55}$ \\
& HAMIL \cite{zhong2023hamil} & TMI'23 & 120.63 & 136.63 &
66.33$_{0.72}$ & 43.07$_{1.18}$ & 33.78$_{1.12}$ &
75.02$_{0.96}$ & 46.10$_{0.32}$ & 22.14$_{0.84}$ \\
& MLPS \cite{han2022multi} & MedIA'22 & 416.64 & 39.56 &
68.14$_{1.03}$ & 46.84$_{0.83}$ & 32.10$_{0.88}$ &
74.11$_{1.38}$ & 45.96$_{0.97}$ & 23.47$_{0.91}$ \\
& SIPE \cite{chen2022self} & CVPR'22 & 611.04 & 26.97 &
60.51$_{0.96}$ & 36.71$_{0.65}$ & 39.24$_{1.45}$ &
70.55$_{0.38}$ & 41.01$_{0.57}$ & 26.89$_{1.10}$ \\
& ARML \cite{feng2024mining} & MICCAI'24 & 403.66 & 40.83 &
68.57$_{1.21}$ & 43.33$_{0.87}$ & 30.94$_{0.96}$ &
76.24$_{1.05}$ & 45.26$_{0.77}$ & 21.08$_{0.68}$ \\
& OEEM \cite{li2022online} & MICCAI'22 & \underline{105.02} & \underline{156.94} &
67.39$_{1.58}$ & 45.08$_{0.73}$ & 32.85$_{1.20}$ &
74.59$_{1.69}$ & 44.17$_{1.12}$ & 22.73$_{0.85}$ \\
& ESFAN \cite{zhang2025edge} & MICCAI'25 & 154.43 & 147.08 &
\underline{71.41$_{0.26}$} & 47.07$_{0.63}$ & \underline{28.08$_{0.48}$} &
\underline{79.29$_{0.35}$} & \underline{49.02$_{0.67}$} & \underline{18.55$_{0.52}$} \\
\rowcolor{gray!10}
& \textbf{Ours} & -- & 191.47 & 86.08 &
\textbf{72.17$_{0.83}$} & \textbf{49.52$_{0.61}$} & \textbf{27.31$_{0.42}$} &
\textbf{79.62$_{0.77}$} & \textbf{49.47$_{0.58}$} & \textbf{18.07$_{0.39}$} \\
\bottomrule
\end{tabular}
}
\end{table*}
\begin{figure}[!t]
    \centering
    \includegraphics[width=\textwidth]{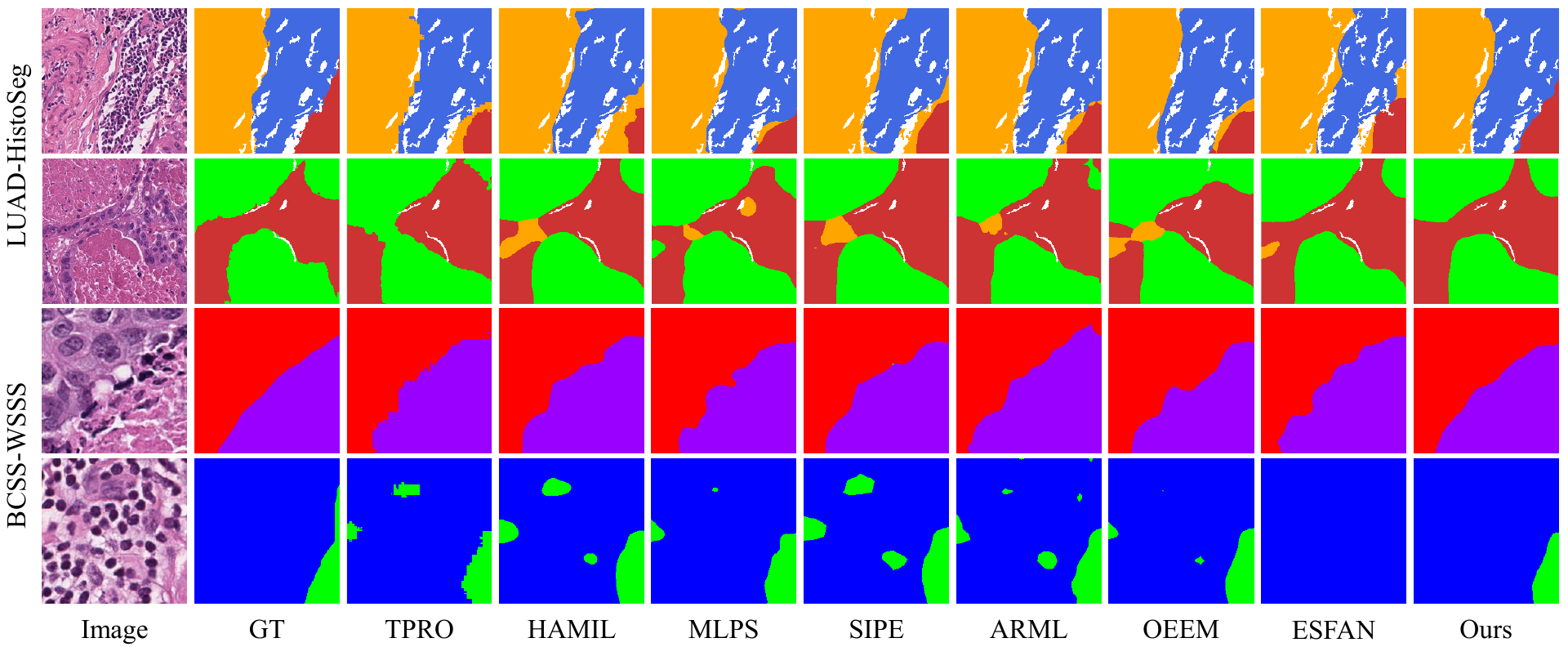}
    \caption{Qualitative segmentation comparisons on the LUAD-HistoSeg and BCSS-WSSS datasets. 
\textbf{Color legend:} LUAD-HistoSeg ( \textcolor{red}{TE}, \textcolor{green}{NEC}, \textcolor{blue}{LYM}, \textcolor{orange}{TAS} ); 
BCSS-WSSS ( \textcolor{red}{TUM}, \textcolor{green}{STR}, \textcolor{blue}{LYM}, \textcolor{violet}{NEC} ).}
\label{fig:qualitative}
\end{figure}
\subsection{Comparison Study} 
As reported in Table~\ref{tab:comparison_subset}, C\textsuperscript{2}RM-Seg establishes the state-of-the-art performance on both benchmarks. On BCSS-WSSS, it achieves 72.17\% mIoU and a HD95 of 27.31, surpassing the leading WSSS competitor, ESFAN. For rigorous evaluation, foundation models (e.g., SAM2, MedSAM) are prompted using our first-stage pseudo-masks; even under this strict condition, our end-to-end framework exhibits superior accuracy. This robust generalization is similarly observed on LUAD-HistoSeg (79.62\% mIoU, 18.07 HD95). Visual comparisons (Fig.~\ref{fig:qualitative}) corroborate these quantitative gains. Compared to baselines, C\textsuperscript{2}RM-Seg delineates complex tissue interfaces with higher boundary fidelity and significantly suppresses false-positive predictions in morphologically ambiguous regions. Despite integrating causal reasoning and foundation priors, our model maintains a competitive computational footprint (191.47 GFLOPs, 86.08 FPS), confirming its practical scalability for large-scale histopathology analysis.
\begin{table}[t]
\centering
\caption{Ablation study of main modules on LUAD-HistoSeg.}
\label{tab:ablation_luad_full}
\setlength{\tabcolsep}{4pt}
\begin{tabular}{ccc|cc|ccc}
\toprule
C\textsuperscript{2}RM & Dual-Path & UGM 
& GFLOPs\(\downarrow\) & FPS\(\uparrow\)
& mIoU\(\uparrow\) & bIoU\(\uparrow\) & HD95\(\downarrow\) \\
\midrule
\(\times\) & \(\times\) & \(\times\) & 112.35 & 208.42 & 71.45\(_{1.12}\) & 39.23\(_{0.94}\) & 27.84\(_{1.56}\) \\
\checkmark & \(\times\) & \(\times\) & 124.68 & 182.15 & 74.12\(_{0.98}\) & 42.56\(_{0.82}\) & 23.45\(_{1.21}\) \\
\(\times\) & \checkmark & \(\times\) & 148.92 & 134.56 & 73.28\(_{1.05}\) & 41.87\(_{0.88}\) & 24.62\(_{1.34}\) \\
\(\times\) & \(\times\) & \checkmark & 135.24 & 168.37 & 75.89\(_{0.92}\) & 44.75\(_{0.79}\) & 21.36\(_{1.12}\) \\
\midrule
\checkmark & \checkmark & \(\times\) & 161.25 & 118.49 & 76.42\(_{0.87}\) & 45.64\(_{0.75}\) & 20.55\(_{0.95}\) \\
\checkmark & \(\times\) & \checkmark & 147.57 & 152.04 & 78.56\(_{0.81}\) & 48.34\(_{0.64}\) & 19.12\(_{0.78}\) \\
\(\times\) & \checkmark & \checkmark & 171.81 & 102.73 & 77.94\(_{0.84}\) & 47.12\(_{0.71}\) & 19.88\(_{0.82}\) \\
\midrule
\checkmark & \checkmark & \checkmark & \textbf{191.47} & \textbf{86.08} & \textbf{79.62\(_{0.77}\)} & \textbf{49.47\(_{0.58}\)} & \textbf{18.07\(_{0.39}\)} \\
\bottomrule
\end{tabular}
\end{table}
\vspace{-5mm}
\subsection{Ablation Study} 
\begin{figure}[!t]
    \centering
    \includegraphics[width=0.9\textwidth]{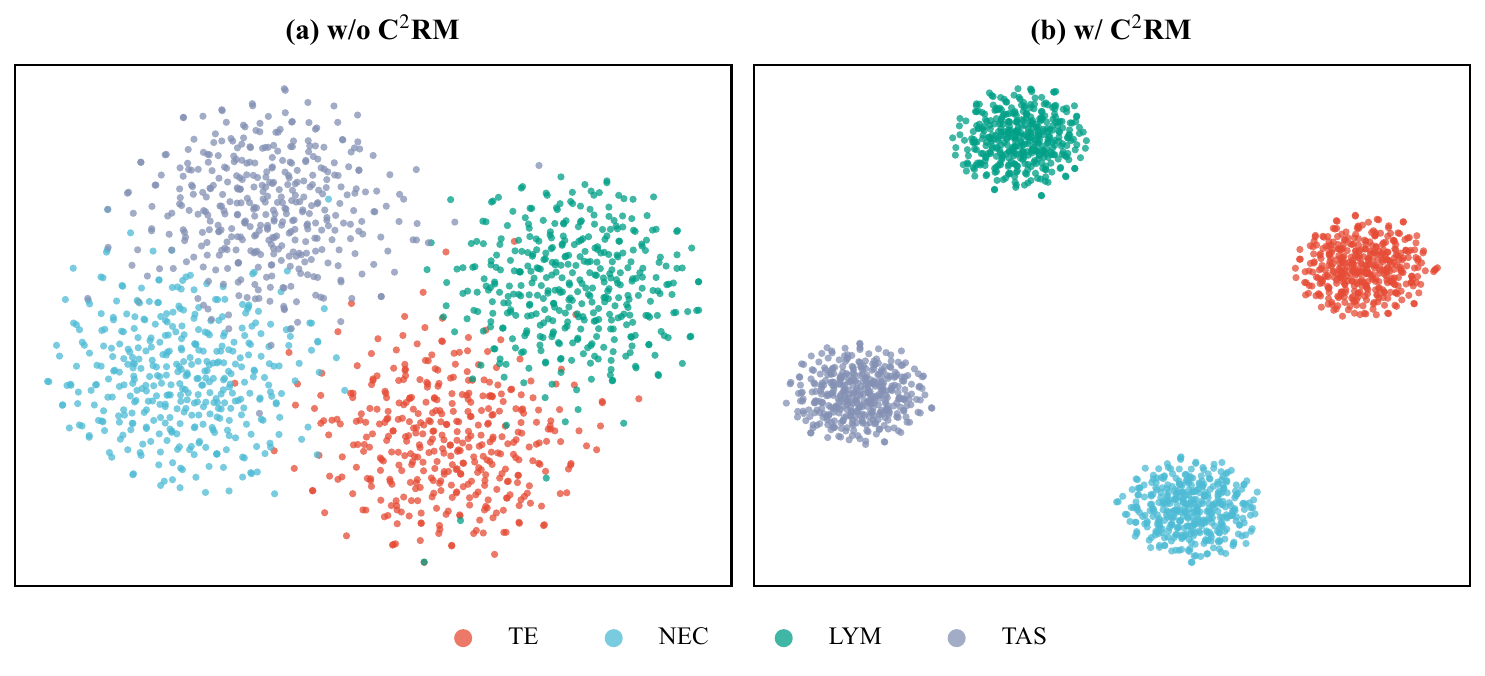}
    \caption{t-SNE visualization. \textbf{(a)} Baseline features are severely entangled by spurious staining confounders. \textbf{(b)} C\textsuperscript{2}RM eliminates these biases via counterfactual intervention, yielding compact clusters and distinct inter-class separability.}
    \label{fig:tsne}
\end{figure}

As shown in Table~\ref{tab:ablation_luad_full}, ablations on LUAD-HistoSeg show that each module independently improves the baseline, indicating complementary corrections of appearance bias, morphology–semantic mismatch, and label noise. In particular, C\textsuperscript{2}RM notably improves both region overlap and boundary quality, suggesting that removing staining-driven confounding restores morphology-aligned pseudo-masks. This deconfounding capability is visually validated by the t-SNE feature space analysis in Fig.~\ref{fig:tsne}, where C\textsuperscript{2}RM transforms severely entangled, stain-biased representations into highly compact and semantically separable clusters. Introducing the Dual-Path architecture further improves boundary fidelity by injecting foundation semantics under structural guidance, reducing morphology–semantic mismatch. The UGM loss significantly decreases HD95, indicating effective suppression of residual pseudo-label noise through uncertainty-aware margin constraints. When combining two modules, performance continues to increase, confirming that causal pseudo-label correction, structure-guided semantic fusion, and noise-aware optimization act synergistically. The full model achieves the best results across all metrics (79.62\% mIoU, 49.47\% bIoU, 18.07 HD95), demonstrating that correcting appearance–morphology bias and enforcing structure-consistent semantics jointly drive the overall performance gain. Despite increased cost, the full model remains computationally efficient.
\section{Conclusion} 
In this paper, we present C\textsuperscript{2}RM-Seg, a novel framework for weakly supervised histopathology image segmentation that fundamentally addresses the pervasive issue of histological confounding. Recognizing that traditional CAM-based pseudo-labels are inherently biased by spurious staining-driven correlations rather than causal tissue morphology, we formulate CAM generation within a rigorous Structural Causal Model. By explicitly executing counterfactual interventions via C\textsuperscript{2}RM, our approach disentangles these confounding factors, synthesizing high-fidelity, morphologically faithful pseudo-masks. To fully leverage these causal priors, we further propose a synergistic Dual-Path   Structural--Semantic Architecture that harmonizes task-specific structural features with frozen foundation-model representations via adaptive gating. Optimized with the UGM loss to ensure robustness against residual noise, C\textsuperscript{2}RM-Seg establishes new state-of-the-art performance on the BCSS-WSSS and LUAD-HistoSeg benchmarks.
\begin{credits}
\subsubsection{\ackname} This work was supported in part by Guangxi Natural Science Foundation (Nos. 2024GXNSFFA010014, AB25069496), National Natural Science Foundation of China (Nos. 82360356, T2541065), and Innovation Project of Guangxi Graduate Education (No. YCSW2025362, YCSW2026398).

\subsubsection{\discintname}
The authors have no competing interests to declare that are relevant to the content of this article.
\end{credits}
\bibliographystyle{splncs04}
\bibliography{Paper-4147}

\end{document}